\documentclass[11pt,a4paper]{article}
\usepackage[hyperref]{acl2019}
\usepackage{times}
\usepackage{latexsym}

\usepackage{url}

\aclfinalcopy

\usepackage{amsfonts}
\usepackage{bm}
\usepackage{array,amsmath}
\usepackage{dsfont}
\usepackage{graphicx}
\usepackage{algorithm}
\usepackage{algorithmicx}
\usepackage{algpseudocode}
\usepackage{pgf,tikz}
\usepackage{mathrsfs}
\usetikzlibrary{arrows}
\usepackage[utf8]{inputenc}
\usepackage{subcaption}
\usepackage{enumerate}
\usepackage{multirow}

\long\def\eat#1{}
\def\bigsep{0.08cm}
\def\bigsepv{7pt}

\def\figref#1{Figure~\ref{fig:#1}}
\def\figlabel#1{\label{fig:#1}\label{p:#1}}

\def\tabref#1{Table~\ref{tab:#1}}
\def\tablabel#1{\label{tab:#1}\label{p:#1}}
\def\secref#1{Section \ref{sec:#1}}
\def\seclabel#1{\label{sec:#1}}
\def\eqref#1{Eq.~\ref{eqn:#1}}

\newcounter{notecounter}
\newcommand{\enotesoff}{\long\gdef\enote##1##2{}}
\newcommand{\enoteson}{\long\gdef\enote##1##2{{
			\stepcounter{notecounter}
			{\large\bf
				\hspace{1cm}\arabic{notecounter} $<<<$ ##1: ##2
				$>>>$\hspace{1cm}}}}}
\enoteson
\enotesoff

\def\method{Co+Co}

\def\MINL{MinLg.}
\def\MAXN{MaxNgr.}

\def\ITER{NIter.}
\def\MINC{MinCount.}
\def\DIM{Dim.}

\def\KJV{James-Ed.}
\def\CAT{Cath-Ed.}
\def\WLD{World-Ed.}
\newcommand{\abbr}[1]{\emph{#1}}

\title{Multilingual Embeddings Jointly Induced from
	Contexts and Concepts:\\
	Simple, Strong and Scalable}

\author{Philipp Dufter, Mengjie Zhao, Hinrich Sch\"{u}tze\\
	Center for Information and Language Processing (CIS), LMU Munich, Germany\\
	{\tt philipp@cis.lmu.de, mengjie.zhao@cis.lmu.de}}

\date{}

\begin{document}
\maketitle

\begin{abstract}
	Word embeddings induced from \emph{local context} are prevalent in NLP.
	A simple and effective context-based multilingual embedding learner is
	\citet{levy2017strong}'s  S-ID (sentence ID)
	method. 
	Another line of work induces
	high-performing
	multilingual embeddings from
	\emph{concepts} \cite{dufter2018embedding}.
	In this paper, we propose
	\method{}, a simple and scalable method that combines
	context-based and concept-based learning.
	From a sentence-aligned corpus,
	concepts are extracted via 
	sampling; words are then associated with
	their concept ID and sentence ID in embedding learning.
	This is the first
	work that successfully combines context-based and
	concept-based embedding learning. We show that \method{}
	performs well for two
	different application scenarios:
	the Parallel Bible Corpus (1000+ languages, low-resource) and EuroParl (12
	languages, high-resource). Among methods  applicable 
	to both corpora, \method{} performs best in our evaluation setup of six tasks.
\end{abstract}

\section{Introduction}

Multilingual embeddings
are useful 
because they provide word 
representations of source and
target language in the same space in machine translation and because they are a basis 
for transfer learning. In contrast to prior multilingual work
\cite{zeman08crosslanguage,mcdonald11delexicalized,tsvetkov14metaphor},
automatically learned embeddings potentially perform as well but are more
efficient and easier to use
\cite{klementiev2012inducing,hermann2014multilingual,guo16transfer}.  Thus,
multilingual word embedding learning is important for natural language processing (\abbr{NLP}).

The quality of multilingual embeddings is driven
by the underlying feature set more than the type of algorithm used for training
the embeddings \cite{upadhyay2016cross,ruder2019survey}.  Most embedding
learners build on using \emph{context information} as feature.
\citet{dufter2018embedding} showed that using \emph{concept
	information} is effective for multilingual embedding learning, as well.
We propose \abbr{\method{}}, a method that 
combines the concept identification method \abbr{Anymalign}
\cite{lardilleux2009sampling} and the multilingual embedding learning method sentence ID (\abbr{S-ID})
\cite{levy2017strong} into a
embedding learning method that is based on both concept
and context.

Our aim is to create a method for learning non-contextualized embeddings that yield strong results while being scalable and widely applicable. Thus we work on two parallel corpora: a low-resource, massively multilingual corpus, the Parallel Bible Corpus (\abbr{PBC}), and on a high-resource, mildly multilingual corpus, \abbr{EuroParl}. Out of 15 embedding learning methods we identify S-ID, the concept based method N(t) by \newcite{dufter2018embedding} and \method{} as the only ones that yield high-quality word spaces across corpora. \method{} exhibits the best and most stable performance.

Our contributions are:
\textbf{i)} We show that Co+Co, i.e.,
using concepts and contexts
jointly,
yields higher quality embeddings than either by
itself.  
\textbf{ii)} We demonstrate that Co+Co performs well across very different datasets and scales across a high number of languages. 
\textbf{iii)} We find
that lower embedding dimensionality is better for word
translation in PBC. In addition, we find that QVEC
evaluation \cite{tsvetkov2015evaluation} is highly dependent on dimensionality.

\section{Methods}
\seclabel{methods}

Our proposed method consists of three steps: 1) Inducing concepts using Anymalign \cite{lardilleux2009sampling}. 2) Generating an artificial corpus using sentence and concept IDs. 3) Training word2vec on the artificial corpus.

\subsection{Concept Induction}

\citet{lardilleux2009sampling} propose Anymalign, an algorithm originally intended for obtaining word alignments. Consider a parallel corpus $V$ across multiple languages.
The central idea is that words that occur strictly in the same sentences, can
be considered translations. In addition to
words, word ngrams can be considered. We call words or word ngrams that occur exclusively in the same
sentences \emph{perfectly aligned}. By this strict definition, the number of perfect alignments is low.
Coverage can be increased by sampling subcorpora.  As the number of
sentences is smaller in each sample and there is a high number of sampled
subcorpora, perfect alignments occur more often.

\figref{anymalign} shows \citet{lardilleux2009sampling}'s Anymalign algorithm on a high level.

There are three relevant hyperparameters in Anymalign: the minimum number of languages a perfect
alignment should cover (\abbr{\MINL{}})\ and the maximum ngram length (\abbr{\MAXN{}}).  The size of
a subsample is adjusted automatically to maximize  the probability that each
sentence is sampled at least once. This probability depends on the
number of  samples drawn and thus on the runtime (\abbr{T}) of the algorithm. Thus,
T is another hyperparameter. For details see
\cite{lardilleux2009sampling}.

We argue that with small \MAXN{} one can interpret perfect alignments as concepts, i.e., a set of semantically similar words.  For example the English trigram ``mount of
olives'' and the French trigram ``montagne des oliviers'' are a perfect alignment describing the same concept.  Thus we define a \emph{concept} as a set of
perfectly aligned words and use Anymalign as concept induction algorithm. Note that most members of a concept are not perfect
alignments in $V$ (only in a subsample $V'$) and that a word can be part of multiple concepts. See \secref{concepts} for comments on concept quality.

\begin{figure}[t]
	\begin{algorithm}[H]
		\scriptsize
		\algrenewcommand\algorithmicindent{0.4cm}
				\caption{\footnotesize Anymalign \cite{lardilleux2009sampling}}
		\begin{algorithmic}[1]
			\Procedure{GetConcepts}{$V$, \MINL{}, \MAXN{}, T}
			\State $C = \emptyset$
			\While {$\textit{runtime} \leq T$}
			\State $V'= \text{get-subsample}(V)$
			\State $A = \text{get-concepts}(V')$
			\State $A = \text{filter-concepts}(A, \text{\MINL{}}, \text{\MAXN{}})$
			\State $C = C \cup A$
			\EndWhile
			\EndProcedure
		\end{algorithmic}
	\end{algorithm}
	\caption{$V$ is a parallel corpus. \emph{get-subsample} creates a sentence-aligned parallel subcorpus by sampling
		lines from $V$. \emph{get-concepts}
		returns the set of perfect alignments (concepts). \emph{filter-concepts}
		filters the set of concepts to enforce \MINL{}\ and \MAXN{} \figlabel{anymalign}}
\end{figure}

\subsection{Corpus Creation}
\seclabel{corpuscreation}

We use sentence IDs from the parallel corpus and concepts to create corpora. \figref{excorpus} shows samples of the
generated corpora to be processed by the embedding learner.

\textbf{S-ID.} We adopt \citet{levy14neural2}'s framework; it formalizes the basic information
that is passed to the embedding learner as a set of pairs. In the monolingual
case, each pair consists of two words that occur in the same context.  A
successful approach to multilingual embedding learning for parallel corpora is
to use  pairs of a word and a sentence ID
\cite{levy2017strong}. The sentence ID acts as crosslingual signal. We call this method \emph{S-ID}.\footnote{Note that we use the name S-ID for the sentence
	identifier, for the corpus creation method that is based on
	these identifiers, for the embedding learning method based
	on such corpora and for the embeddings produced by the
	method. The same applies to other method names. Which sense is meant
	should be clear from context.}

\textbf{C-ID.} We propose to adjust S-ID by replacing sentence IDs with concept IDs. That is, a line consists of a concept ID and a concept member word.

\textbf{\method{}.} We combine S-IDs and C-IDs by creating two corpora with the respective method and then concatenating their corpora before
learning embeddings. Intuitively, both methods use complementary information: S-ID uses local context information and C-ID leverages globally aggregated information. Therefore, we expect a higher performance by combining those.

\begin{figure}[t]
	\scriptsize
	\def\bigsep{0.1cm}
	\begin{tabular}{c}
		{\begin{tabular}{l@{\hspace{\bigsep}}l@{\hspace{\bigsep}}}
				\multicolumn{2}{c}{[...]}\\
				\texttt{48001018}& \texttt{enge:fifteen}\\
				\texttt{48001018}& \texttt{enge:,}\\  
				\texttt{48001018}& \texttt{enge:years}\\
				\texttt{48001018}& \texttt{deu0:fünfzehn}\\
				\texttt{48001018}& \texttt{deu0:Jahre}\\
				\texttt{48001018}& \texttt{deu0:,}\\
				\multicolumn{2}{c}{[...]}
		\end{tabular}}
		{\begin{tabular}{l@{\hspace{\bigsep}}l@{\hspace{\bigsep}}}
				\multicolumn{2}{c}{[...]}\\
				\texttt{C:911} & \texttt{kqc0:Jerusalem} \\
				\texttt{C:911} & \texttt{por5:Jerusalém} \\
				\texttt{C:911} & \texttt{eng7:Jerusalem} \\
				\texttt{C:911} & \texttt{haw0:Ierusalema} \\
				\texttt{C:911} & \texttt{ilb0:Jelusalemu} \\
				\texttt{C:911} & \texttt{fra1:Jérusalem} \\
				\multicolumn{2}{c}{[...]}
		\end{tabular}} \\\hline
		{\begin{tabular}{l@{\hspace{\bigsep}}p{5.5cm}@{\hspace{\bigsep}}}
				\multicolumn{2}{c}{[...]}\\
				\texttt{45016016} & Salute one another with an holy kiss . The churches of Christ salute you . \\
				\texttt{48001018} & Then after three years I went up to Jerusalem to see Peter  ,
				and abode with him fifteen days . \\
				\multicolumn{2}{c}{[...]}
		\end{tabular}} 
	\end{tabular}
	
	\caption{Samples of S-ID (top-left) and C-ID (top-right) corpora that
		are input to word2vec. Words are prefixed by a 3
		character ISO 639-3 language identifier followed by
		an alphanumeric character to distinguish 
		editions in the same language (e.g., enge =
		English King James Version (\KJV{})). C:911 is a concept identifier.
		Bottom: \KJV{} text
		sample.
		\figlabel{excorpus}}
\end{figure}

\subsection{Embedding Learning}
\seclabel{embeddinglearning}

To obtain embeddings we use use \emph{word2vec skipgram}\footnote{We use {
		code.google.com/archive/p/word2vec}} \cite{mikolov2013efficient} on the 
	generated corpora. We use default
hyperparameters,  and investigate three more closely:  number of iterations (\abbr{\ITER{}}), minimum
frequency of a word (\abbr{\MINC{}})\ and embedding dimensionality (\abbr{\DIM{}}). Throughout the paper we $\ell^2$-normalize vectors.

\section{Application to Parallel Bible Corpus}
\subsection{Data}

We work on \abbr{PBC}, the Parallel Bible Corpus \cite{mayer2014creating}, a
verse-aligned corpus of 1000+ translations of the New Testament.  For the sake
of comparability we use the same 1664 Bible editions across 1259 languages
(distinct ISO 639-3 codes) and the same 6458 training verses as in
\cite{dufter2018embedding}.\footnote{ Information downloaded from
	cistern.cis.lmu.de/comult} We follow their terminology and refer to
``translations'' as ``editions''. PBC
is a good model for resource-poverty; e.g., the training set of the English King James Version (\abbr{\KJV{}})\ contains
fewer than 150,000 tokens in 6458 verses. \KJV{}\ spans a vocabulary of 6162 words and all 32 English editions together cover 23,772 words. 
We use the tokenization provided in the data, which is erroneous for
some hard-to-tokenize languages (e.g., Khmer, Japanese), and do not apply
further preprocessing.

\subsection{Evaluation}

\citet{dufter2018embedding} introduce roundtrip translation (\abbr{RTT}) as multilingual
embedding evaluation when no gold standard is available. 
A query word $q$ in language $L_1$ is translated to its nearest neighbor (by cosine similarity) $v$ in an intermediate language $L_2$ and then backtranslated to its
closest neighbor $q'$ in the query language $L_1$. RTT is successful if $q=q'$.
For the roundtrips we consider $k_I$ nearest neighbors in $L_2$
and, for each of these intermediate neighbor, $k_T$
predictions in $L_1$. 

Predictions are compared to a ground truth set $G_q$. There is a strict ($\{q\}$) and
a relaxed (\{words with the same lemma and part-of-speech as $q$\}) ground truth. Thus multiple roundtrips of $q$ can be considered correct (e.g., inflections of a query). We average binary
results (per query) over editions and report mean ($\mu$)
and median (\abbr{Md.)}\ over queries. Inspired by the precision@k evaluation for
word translation we vary $(k_I, k_T)$ as follows: ``S1'' $(1,1)$, ``R1''
$(1,1)$, ``S4'' $(2,2)$, and ``S16'' $(2,8)$, where S (R) stand for using the strict (relaxed) ground truth.  

If $q$ is not in the embedding space, we consider the roundtrip as failed. 
The number of queries contained in the embedding space is denoted by $N$ (``coverage''). We use the same queries as in
\cite{dufter2018embedding}, which are based on \cite{swadesh1971south}'s 100
universal words. In addition, we introduce
a development set: an earlier list by Swadesh with
215 words.\footnote{concepticon.clld.org/contributions/Swadesh-1950-215}  151 queries remain in the development set after
excluding queries from the test set.
Due to this large number we do not compute the relaxed measure on the development set as this requires
manual effort on the ground truth.
We work on the level of Bible editions, i.e., two editions in the same language
are considered different ``languages''.	
We use \KJV{}\ as the query edition if \KJV{}\ contains
$q$. Else we
randomly choose another English edition.

As extrinsic task we perform sentiment analysis following \cite{dufter2018embedding}. We use their ground truth and data split into training and test set. There are two classifications: whether a verse contains positive (\abbr{Pos.})\ or negative (\abbr{Neg.})\ sentiment. Given the multilingual space, for each task a linear support vector machine (\abbr{SVM}) is trained on the English World Edition (\abbr{\WLD{}})\ (following \cite{dufter2018embedding}) and subsequently tested on all other editions.
We train SVMs in 5-fold cross-validation on the training set to optimize the hyperparameter $C$. We report average $F_1$ across editions.

\subsection{Baselines}

We compute diverse  baselines  to pinpoint
reasons for performance changes 
as much as possible.

\textbf{Monolingual Embedding Space.}
In the baseline \abbr{MONO} we train 1664 monolingual spaces using
word2vec and interpret them as if they were a shared multilingual embedding space. 
This serves as consistency check for our evaluation methods.

\textbf{Context Based Embedding Space.}
We use
\abbr{S-ID} to obtain a multilingual space.

\textbf{Transformation Based Embedding Space.}
The baseline \abbr{LINEAR}
follows \cite{duong2017multilingual}. We pick one edition as the ``embedding
space defining edition'', in our case the English Catholic Bible (\abbr{\CAT{}}). We did not choose \KJV{}\ or \WLD{}\ to avoid that one single edition has multiple special roles. We then create 1664
bilingual embedding spaces using S-ID; in each case the two editions covered are \CAT{}\
and one of the other 1664 languages.  We then use
\cite{mikolov2013exploiting}'s linear transformation method to map all
embedding spaces to the \CAT{}\ embedding space. More specifically, let $X' \in
\mathbb{R}^{n_X\times d}$, $Y'\in \mathbb{R}^{n_Y\times d}$ be two embedding
spaces, with $n_X$, $n_Y$ number of words, and
$d$ be the embedding dimension. We then select $n_T$ transformation words that
are contained in both embedding spaces. This yields: $X, Y \in \mathbb{R}^{n_T\times d}$.
The transformation matrix $W \in
\mathbb{R}^{d \times d}$ is then given by 
${\arg\min_W} \lVert XW - Y \rVert_F$ where $\lVert \cdot \rVert_F$ is the Frobenius norm.
The closed form solution is given by $W^* = X^+Y $ where $ X^+$
is the Moore-Penrose Pseudoinverse \cite{penrose1956best}. In our case, $X$ is a bilingual and $Y$ is a monolingual embedding space, both containing the vocabulary of \CAT{}

\textbf{Bilingual Embedding Spaces.}
\abbr{BILING}  uses the same bilingual embedding spaces as
LINEAR, but we do not transform the bilingual spaces into a common space. 
This baseline shows the effect of multilingual vs.\  bilingual embedding spaces. As we do not have a multilingual space we need to modify our evaluation methods slightly: we perform RTT
in each bilingual embedding space
separately. For sentiment
analysis, we train one SVM
per embedding
space on English, which is then tested on the other edition.

\textbf{Unsupervised Embedding Learning.}
We apply the recent unsupervised embedding learning method by
\citet{lample2018word} (\abbr{MUSE}).\footnote{We use https://github.com/facebookresearch/MUSE} Given
unaligned corpora in two languages,
MUSE
learns two separate embedding spaces that are subsequently unified 
by
a linear transformation. This transformation is learned using a discriminator neural network
that tries to identify the original language of a vector. Again, we learn monolingual embedding spaces by running word2vec on PBC directly. Subsequently, we transform all spaces into the word space of \CAT{}\ using MUSE.
\citet{chen2018unsupervised} extended MUSE multilingually. We include their method \abbr{MAT+MPSR} as baseline. MAT+MPSR is memory and computation intensive. Allocating three days of computation on a standard GPU (GTX 1080 Ti), we were only able to apply this baseline on a subset of 52 editions.

\textbf{Non-Embedding Baseline.} 
To show that embedding spaces  provide
some advantages over using the concepts as is, we
introduce
\abbr{C-SIMPLE}, a non-embedding baseline that follows the idea of
RTT. Given a query word $q$ and an intermediate edition, we
consider all words that share a concept ID with $q$ as possible
intermediate words. We then choose randomly (probability weights
according to number of concepts shared with $q$) intermediate words. For
back translation we apply the same procedure.

\begin{table}[t]
	\centering
	\scriptsize
		\begin{tabular}{l@{\hspace{\bigsep}}r@{\hspace{\bigsep}}r@{\hspace{\bigsep}}r@{\hspace{\bigsep}}r@{\hspace{\bigsep}}||r@{\hspace{\bigsep}}r@{\hspace{\bigsep}}|r@{\hspace{\bigsep}}r@{\hspace{\bigsep}}|r@{\hspace{\bigsep}}r@{\hspace{\bigsep}}|r@{\hspace{\bigsep}}}
		& &&&& \multicolumn{2}{c}{S1} &  \multicolumn{2}{c}{S4} & \multicolumn{2}{c}{S16}&\\
		&  &  \rotatebox{90}{\ITER{}}&\rotatebox{90}{\MINC{}}&\rotatebox{90}{\DIM{}}&$\mu$ & Md & $\mu$ & Md &$\mu$ & Md  & N\\
		\hline \hline
		\rule{0pt}{\bigsepv}1 & S-ID & \textbf{100}&5&200 & 29 & 21 & 43 & 46 & 56 & 78 & 103\\
		\hline
		\rule{0pt}{\bigsepv}2 & S-ID & 5&& & 14 & 11 & 25 & 22 & 41 & 45 & 103\\
		3 & S-ID & 10&& & 25 & 16 & 38 & 34 & 51 & 60 & 103\\
		4 & S-ID & 25&& & 27 & 20 & 41 & 43 & 53 & 69 & 103\\
		5 & S-ID & 50&& & 27 & 16 & 40 & 40 & 53 & 67 & 103\\
		6 & S-ID & 150&& & 29 & 21 & 43 & 47 & 56 & 79 & 103\\
		7 & S-ID & &\textbf{2}& & \textbf{35}  & \textbf{31}  & \textbf{52}  & \textbf{60}  & \textbf{66}  & \textbf{90}  & 130\\
		8 & S-ID & &10& & 24 & 5 & 36 & 17 & 48 & 63 & 85\\
		9 & S-ID & &&\textbf{100} & 30 & 24 & 45 & 48 & 58 & 83 & 103\\
		10 & S-ID & &&300 & 28 & 19 & 42 & 41 & 54 & 69 & 103\\
	\end{tabular}
	\\
	\vspace{0.2cm}
	\centering
		\begin{tabular}{l@{\hspace{\bigsep}}r@{\hspace{\bigsep}}r@{\hspace{\bigsep}}r@{\hspace{\bigsep}}r@{\hspace{\bigsep}}||r@{\hspace{\bigsep}}r@{\hspace{\bigsep}}|r@{\hspace{\bigsep}}r@{\hspace{\bigsep}}|r@{\hspace{\bigsep}}r@{\hspace{\bigsep}}|r@{\hspace{\bigsep}}}
		& &&&& \multicolumn{2}{c}{S1} & \multicolumn{2}{c}{S4} & \multicolumn{2}{c}{S16}&\\
		&  &  \rotatebox{90}{\MINL{}}&\rotatebox{90}{\MAXN{}}&\rotatebox{90}{T}&$\mu$ & Md & $\mu$ & Md &$\mu$ & Md  & N\\
		\hline \hline
		\rule{0pt}{\bigsepv}1 & C-ID & \textbf{100}&\textbf{3}&10 & \textbf{30}  & \textbf{26}  &  \textbf{46}  & \textbf{46}  & 60 & 71 & 120\\
		\hline
		\rule{0pt}{\bigsepv}2 & C-ID & 50&& & 26 & 21 &  39 & 42 & 56 & \textbf{79}  & 104\\
		3 & C-ID & 150&& & 22 & 14 &  34 & 28 & 47 & 56 & 94\\
		4 & C-ID & 500&& & 15 & 0 &  25 & 0 & 33 & 0 & 59\\
		5 & C-ID & &1& & 17 & 0 &  27 & 0 & 38 & 0 & 73\\
		6 & C-ID & &5& & 28 & 20 & 40 & 35 & 55 & 58 & 122\\
		7 & C-ID & &&5 & 24 & 20 & 36 & 34 & 48 & 52 & 114\\
		8 & C-ID & &&\textbf{15} & \textbf{30}  & \textbf{26}  &  \textbf{46}  & 44 & \textbf{61}  & 76 & 121\\
	\end{tabular}
	\caption{Hyperparameter selection for word2vec
		(top) and Anymalign (bottom) on RTT. Initial parameters
		in first row; empty cell:
		initial parameter from the first row. For example: to compare the effect of 
		different dimensionality in word2vec compare lines 1, 9 and 10 in the top table (best dimension is 100). Bold: best result per column or selected
		hyperparameter value.}
	\tablabel{hyper}
\end{table}

\subsection{Hyperparameter Selection}\seclabel{hyperparams}\seclabel{exp}

We select hyperparameters based on the roundtrip translation task.

\textbf{Word2vec.}  \seclabel{ablationword2vec}
We tune word2vec parameters based on the method S-ID.
Since a grid search for optimal values for
the parameters \ITER{},\ \MINC{}\ and \DIM{}\
would take too long, we search greedily instead: we choose an initial parameter setting, vary one parameter at a time and select the value with the best performance.
More iterations
yield better performance. \ITER{}\ = 100 is a good
efficiency-performance trade-off.  For \MINC{}\ (minimum
frequency of a word) the best performance is found using 2. This is mainly 
due to increased coverage. 
Surprisingly,
smaller embedding dimensions
work better to
some degree.
A highly multilingual embedding space is expected to suffer
more from ambiguity and that is an argument for higher dimensionality;
cf.\ \citet{li15multisense}. But this effect seems to be
counteracted by the low-resource properties of PBC
for which the increased number of parameters of higher
dimensionalities cannot be estimated reliably.
We choose embedding size 100.

\textbf{Anymalign.}
We
tune the hyperparameters 
for Anymalign by evaluation embedding spaces created with C-ID. We use the above word2vec settings except for \MINC{} As argued before, each word in a concept carries a strong multilingual signal, which 
is why we do not apply any frequency filtering for
C-ID. Thus we set \MINC{}\ to 1 whenever we learn concept based
embeddings. For \method{} we apply the different frequency
thresholds on the two parts (S-ID and C-ID parts)
of the corpus.

We
find the best performance when setting the minimum number of editions (\MINL{})\ to 100. 
As expected coverage worsens when increasing \MINL{}
\MAXN{}\ is the maximum ngram length.
We see the best performance when setting \MAXN{}\ to 3.
This is intuitive for 
languages like Swedish since  compounds can be found.
T is 
the time in hours, i.e.,  for how long to
sample (which steers the size and number of sampled subcorpora).
As expcted higher T yields better performance. Given only a slight difference between 10 and 15 
we set T to 15 for slightly higher coverage.

\begin{table} \centering
	\scriptsize
	\def\bigsep{0.06cm}
	\begin{tabular}{l@{\hspace{\bigsep}}r@{\hspace{\bigsep}}||r@{\hspace{\bigsep}}r@{\hspace{\bigsep}}|r@{\hspace{\bigsep}}r@{\hspace{\bigsep}}|r@{\hspace{\bigsep}}r@{\hspace{\bigsep}}|r@{\hspace{\bigsep}}r@{\hspace{\bigsep}}|r@{\hspace{\bigsep}}||r@{\hspace{\bigsep}}|r@{\hspace{\bigsep}}}
		& & \multicolumn{9}{c}{Roundtrip} & \multicolumn{2}{c}{Sentiment}\\
		& & \multicolumn{2}{c}{S1} & \multicolumn{2}{c}{R1}  & \multicolumn{2}{c}{S4} &
		\multicolumn{2}{c}{S16}&&&\\
		&  & $\mu$ & Md & $\mu$ & Md& $\mu$ & Md  &$\mu$ &
		Md  & N &Pos. & Neg. \\
		\hline \hline 
		\rule{0pt}{\bigsepv}1 & C-SIMPLE & 35 & 33 & 35 & 34 & 49 & 54 & 56 & 56 & 67&&\\
		2 & MONO & 21 & 18 & 21 & 18 & 39 & 36 & 67 & 68 & 69&4&74\\
		3 & S-ID & 48 & 47 & 53 & 59 & 65 & 72 & 83 & 93 & 69 & 71 & \underline{88}\\
		4 & C-ID & 43 & 42 & 46 & 43 & 58 & 60 & 79 & 91 & 67 & 73 & 86\\
		5 & \method{} & \underline{51} & \underline{50} & \underline{56} & \underline{65} & \underline{69} & \underline{80} & \underline{86} & \underline{96} & 69 & 79 & \textbf{89}\\
		6 & LINEAR & 30 & 28 & 31 & 30 & 43 & 46 & 62 & 71 & 69 & 76 & 83\\
		7 & MUSE & 16 & 13 & 16 & 13 & 33 & 28 & 65 & 64 & 69 & 13 & 73\\
		8 & MAT+MPSR & 11 & 10 & 11 & 10 & 20 & 17 & 44 & 42 & 66 & 37 & 72\\
				9 & BILING & 41 & 32 & 42 & 35 & 53 & 55 & 67 & 82 & 69&\underline{80}&87\\
		10 & N(t)* & \textbf{54}  & \textbf{59}  & \textbf{61}  & \textbf{69}  & \textbf{80}  & \textbf{87}  & \textbf{94}  & \textbf{100}  & 69 & \textbf{82} & \textbf{89}\\
		11 & SAMPLE* & 33 & 23 & 43 & 42 & 54 & 59 & 82 & \underline{96} & 65 & \textbf{82} & \textbf{89}\\
	\end{tabular} 
	\caption{Evaluation results.
		*: results by \cite{dufter2018embedding}. Sentiment analysis not defined for C-Simple. Bold: best result per column, Underlined: second best.} \tablabel{bigtable}
\end{table}

\subsection{Results}
\tabref{bigtable} presents results on the test set.

\textbf{Embeddings vs.\  Concepts.} 
C-Simple works reasonably well but is outperformed by most
embedding spaces. This indicates that learning embeddings
augments the concept information.

\textbf{Evaluation Difficulty.} 
MONO is the weakest baseline
and is a good indicator that
the RTT task is 
challenging, e.g.,  S1=21 and
R1=18. It admits the weakness of RTT by yielding a non-zero performance
despite most of the
intermediate neighbours being not related 
to the query at all. Still it proves that RTT is
a good indicator for the
quality of multilingual word
spaces, as truly
multilingual word spaces
significantly outperform
MONO (especially
for the median). 
The same is
true for
sentiment analysis where
Pos.\ is more challenging
than Neg. Thus Pos.\ is a better
indicator of 
performance differences.

\textbf{Transformation Based Spaces.} 
LINEAR
performs similar to C-Simple (but outperforms it for S16). This supports the hypothesis that PBC offers too little
data for training mono-/bilingual embeddings. As
expected BILING works better than LINEAR.  Keep in mind that this is not a universal embedding space
and has fewer constraints than other embedding spaces. Thus it
is not directly comparable.

\textbf{Unsupervised Embedding Learning.} 
MUSE performs even worse than MONO. MAT+MPSR performs poorly,
as well. This is a strong indication that PBC offers too
little data for learning high-quality monolingual
embeddings. In addition, we hypothesize that the word spaces
themselves offer too few data points (i.e.,\ vocabulary
size) for neural network based mapping approaches. We plan
to investigate this  in the
future. As discussed
MAT+MPSR results are for a subset of 52 editions.

\textbf{Context vs.\  Concept Based.} 
N(t) by \citet{dufter2018embedding} shows consistently the
best performance. Later we
will see that this method works
well
here (massively
multilingual PBC setting),
but worse on EuroParl. \abbr{SAMPLE} by
\citet{dufter2018embedding} is based on sampling like
C-ID. However, it
induces concepts only on a small set of pivot languages, not
on all
1664 editions. This does not only work worse (C-ID beats
SAMPLE  except for S16), but it also requires word alignment
information and is thus computationally more expensive. 
As has been observed
frequently, S-ID is highly
effective. S-ID ranks third consistently. Representing a word
as its binary vector of verse occurrence 
provides a clear signal to the embedding learner. Concept-based methods can be considered
complementary to this feature set because they can exploit aggregated
information across languages as well as across verses.
C-ID alone has slightly lower performance than S-ID.

\textbf{Combining Concept and Context.}
\method{}, the combination, outperforms both C-ID and S-ID and
seems to unite the ``best of both worlds''. It yields
consistently the best performance with 3\% to 6\% relative performance increase (for $\mu$) compared to S-ID alone and even more compared to C-ID. Overall \method{}
always ranks second. Thus
combining context and
concept is effective indeed,
but not sufficient to
outperform the strong method
N(t) which is tailored for
massively multilingual
corpora. In experiments we
found that adding C-ID to
N(t) harmed the performance
of N(t) severely, so these
two methods seem to be incompatible.
In short: N(t) is the best method on
this corpus,   \method{} second-best and S-ID is third.

\section{Application to a High-Resource Corpus}

PBC is a low-resource, highly multilingual scenario. We now
provide experimental evidence that \method{} is broadly applicable
and works
in a high-resource, mildly multilingual scenario.
We test the three best performing methods (based on PBC S1 $\mu$):
S-ID, \method{}, N(t).

\subsection{Data}
We choose a dataset by
\citet{ammar2016massively}, a parallel corpus covering
12 languages\footnote{Bulgarian,
	Czech, 
	Danish, 
	English, 
	Finnish, 
	French, 
	German, 
	Greek, 
	Hungarian, 
	Italian, 
	Spanish, 
	Swedish} from the proceedings of the European
Parliament, Wikipedia titles and news commentary.
We refer to this corpus as \abbr{EuroParl}. We do not apply any preprocessing. The dataset is lowercased.

\subsection{Evaluation}
\citet{ammar2016massively} provide an extensive evaluation framework
covering two extrinsic and four intrinsic tasks. The
tasks are document classification, dependency parsing (those models were not available to us, so we omit this evaluation), word translation, word
similarity, \abbr{QVEC}  \cite{tsvetkov2015evaluation} and \abbr{QVEC-cca}
\cite{ammar2016massively}. For all tasks, there is a development and test set available. 
For more details on  data and tasks see \cite{ammar2016massively}.
Due to obvious weaknesses of QVEC (measure not rotation invariant, for details see \cite{ammar2016massively} and \tabref{eurodim}) we omit QVEC in our final evaluation. In their word translation task \citet{ammar2016massively} reduce the word space to contain only words from the evaluation test set. We are interested in an (unrestricted) word translation task, where all words in the word space are possible answers, which is why we reimplemented this task and report results (precision@1) only for the unrestricted word translation task. Obviously this task is more challenging and thus the performance numbers we report are significantly lower than the numbers reported by \newcite{ammar2016massively}.

To ensure comparability with previous approaches, we follow \citet{ammar2016massively} 
in evaluating only on words that are contained in the embedding space and simultaneously reporting the coverage 
(e.g., how many queries of the task are contained in the
embedding space). Only for word translation we follow the same reasoning as for PBC and compute accuracy across all queries (i.e.,  queries not in the word space count as errors).

\def\ephypsep{0.05cm}

\begin{table}[t]
	\scriptsize
	\centering
	\begin{tabular}{c@{\hspace{\ephypsep}}|c@{\hspace{\ephypsep}}|c@{\hspace{\ephypsep}}}
		\method{} & S-ID & N(t)\\\hline
		\begin{tabular}{r@{\hspace{\ephypsep}}r@{\hspace{\ephypsep}}|r@{\hspace{\ephypsep}}}
			\rotatebox{90}{\MINL{}} & \rotatebox{90}{\MINC{}} &  \rotatebox{90}{\begin{tabular}{l}Word\\ Trans.\end{tabular}} \\
			\hline
			\hline
			\rule{0pt}{\bigsepv} 6 & 5 & 18.85 \textsubscript{52.37} \\
			\hline
			\rule{0pt}{\bigsepv}3 &  &  18.38 \textsubscript{52.09}\\
			\textbf{9} &  & 20.15 \textsubscript{52.37} \\
			12 &  & 19.87 \textsubscript{52.37}\\
			& \textbf{2} & \textbf{21.17} \textsubscript{67.41} \\
			& 10 & 17.92 \textsubscript{50.51} 
		\end{tabular}&
		\begin{tabular}{r@{\hspace{\ephypsep}}|r@{\hspace{\ephypsep}}}
			\rotatebox{90}{\MINC{}} &   \rotatebox{90}{\begin{tabular}{l}Word\\ Trans.\end{tabular}}  \\
			\hline
			\hline
			\rule{0pt}{\bigsepv}5 & 18.48 \textsubscript{52.37} \\
			\hline
			\rule{0pt}{\bigsepv}\textbf{2} &\textbf{19.22} \textsubscript{65.65} \\
			10 & 15.97 \textsubscript{43.92} \\
		\end{tabular}&
		\begin{tabular}{r@{\hspace{\ephypsep}}|r@{\hspace{\ephypsep}}}
			\rotatebox{90}{NPiv.} &  \rotatebox{90}{\begin{tabular}{l}Word\\ Trans.\end{tabular}} \\
			\hline
			\hline
			\rule{0pt}{10pt}
			\textbf{4} &  \textbf{17.08} \textsubscript{59.33}\\
			\hline
			\rule{0pt}{\bigsepv}2 &  15.88 \textsubscript{50.05} \\
			6 & 16.90 \textsubscript{62.95}
		\end{tabular}
	\end{tabular}
	\caption{Hyperparameter
		selection on
		EuroParl. Initial
		parameter in first
		row; empty cell:
		initial parameter
		from first row. Bold: best result per column or selected
		hyperparameter values. Subscript numbers indicate coverage.}
	\tablabel{ephyper}
\end{table}

\subsection{Hyperparameter Selection}

\begin{table}[!t]
	\centering
	\scriptsize
	\def\epsep{0.15cm}
	\begin{tabular}{r@{\hspace{\epsep}}r@{\hspace{\epsep}}||r@{\hspace{\epsep}}|r@{\hspace{\epsep}}|r@{\hspace{\epsep}}|r@{\hspace{\epsep}}|r@{\hspace{\epsep}}}
		&\multirow{2}{\epsep}{\rotatebox{90}{\DIM{}}}& Word & Word &QVEC&QVEC-&  Doc. \\
		&&Trans. & Sim. &  & cca &  Class. \\
		\hline
		\rule{0pt}{\bigsepv}
		S-ID & 10  & 0.19   &23.61   & 12.10   & 19.75   & 53.30  \\
		S-ID & 25  & 4.92   &  42.08   & \textbf{15.74}   & 19.24   & 74.33  \\
		S-ID & 50  & 13.93   &  54.29   & 14.85   & 19.67   & 83.91  \\
		S-ID & 100  & 15.97   & 53.86   & 11.73   & 25.51   & 86.88  \\
		S-ID & 300  & 16.81   &  \textbf{56.29}   & 9.29   & 34.58   & 90.31  \\
		S-ID & 500  & \textbf{16.90}   & 55.51   & 9.01   & \textbf{38.64}   & \textbf{90.88}  \\
		\hline
		\textsubscript{Coverage} && \textsubscript{43.92}  & \textsubscript{57.58}  &  \textsubscript{70.91}  & \textsubscript{70.91}  & \textsubscript{38.89}
	\end{tabular}
	\caption{Results of S-ID on the dev set for varying dimensionality. There is clear correlations with the embedding dimension. Note that results are slightly different to \tabref{ephyper} due to \MINC{}\ 10 used in this table.
	}
	\tablabel{eurodim}
\end{table}

We optimize corpus
specific
hyperparameters  (e.g., \MINL{})\ on the development
set of the word translation task.
\method{}: we vary the minimum number of languages
that need to be covered by the concept identification and \MINC{}\ for the S-ID part of \method{}. N(t) requires pivot
languages. Following \citet{dufter2018embedding} we 
choose as pivot languages those with the lowest type-token ratio
(these are Greek, Danish, Spanish, French, Italian, English) and vary the number of pivot languages (\abbr{NPiv.})\ between 2, 4 and 6.

\tabref{ephyper} gives an overview of our hyperparameter selection. For
\method{}, we choose \MINL{}\ 9 and \MINC{}\ 2. 
For S-ID, we find the best performance with \MINC{}\ 2.  For
N(t), the best result is obtained when using 4 pivot languages.

Further, we show the effect of varying embedding dimensions in \tabref{eurodim}. For most tasks 300 to 500 dimensions are optimal. This confirms the findings by \newcite{yin2018dimensionality}. For QVEC, extremely low dimensionality is beneficial. Note that QVEC is not rotation invariant: we hypothesize that the probability of an axis being highly correlated with linguistic features in a high-dimensional space is very small compared to a low-dimensional space. QVEC-cca on the other hand benefits greatly from higher dimensions and even choosing 100 dimensions, which is a popular and reasonable choice, seriously harms the performance compared to 300 dimensions: the higher the dimensionality the more likely, CCA will find an arbitrary dimension which is highly correlated with linguistic features. When comparing our results to \cite{ammar2016massively} we need to be aware of this effect, as they used an embedding dimension of 512 vs.\  100 used in this work.

\subsection{Results}

\begin{table}[!t]
	\centering
	\scriptsize
	\def\epsep{0.08cm}
	\begin{tabular}{@{\hspace{\epsep}}r@{\hspace{\epsep}}||@{\hspace{\epsep}}r@{\hspace{\epsep}}|@{\hspace{\epsep}}r@{\hspace{\epsep}}|@{\hspace{\epsep}}r@{\hspace{\epsep}}|r@{\hspace{\epsep}}r@{\hspace{\epsep}}r@{\hspace{\epsep}}}
		& Word  & Word &QVEC-& Doc.  \\
		& Trans. & Sim. & cca & Class.\\
		\hline
		\hline
		\rule{0pt}{\bigsepv}
		multiCluster* & 11.79 \textsubscript{62.30}  & 57.45 \textsubscript{73.89}  & 43.34 \textsubscript{82.01}  & 92.11 \textsubscript{48.16} \\
		multiCCA* & 11.79 \textsubscript{77.16}  & \emph{69.99} \textsubscript{77.94}  & 41.52 \textsubscript{87.03}  & \emph{92.18} \textsubscript{62.81} \\
		multiSkip* & 11.70 \textsubscript{54.41}  & 60.24 \textsubscript{67.55}  & 36.34 \textsubscript{75.69}  & 90.46 \textsubscript{45.73} \\
		Invariance* & 12.26 \textsubscript{41.41}  & 59.13 \textsubscript{62.50}  & \emph{46.21} \textsubscript{74.78}  & 91.10 \textsubscript{31.35} \\
		\hline
		S-ID & 19.13 \textsubscript{64.53}  & 48.62 \textsubscript{75.39}  & 24.58 \textsubscript{80.33}  & 86.66 \textsubscript{56.45} \\
		\method{} & \emph{\textbf{20.24}} \textsubscript{65.92}  & \textbf{52.75} \textsubscript{75.39}  & 24.39 \textsubscript{82.76}  & \textbf{87.33} \textsubscript{57.52} \\
		N(t) & 15.32 \textsubscript{59.42}  & 48.01 \textsubscript{71.27}  & \textbf{25.92} \textsubscript{79.07}  & 84.97 \textsubscript{53.17}
	\end{tabular}
	\caption{Results on the test set. *: methods by \cite{ammar2016massively}. We downloaded their embedding spaces
		and performed the evaluation using their
		code.
		We can mostly reproduce their results (up to
		rounding errors), but 
		word similarity
		numbers are slightly different.
		Best result across S-ID, \method{}, N(t) is bold; best result across all methods is italic.
	}
	\tablabel{europarl}
\end{table}

\tabref{europarl} provides results on the test set. It immediately becomes clear that different word spaces have different strengths and weaknesses. 

Among S-ID, \method{} and N(t), \method{} performs best, followed by S-ID and N(t) in 3 out of 4 tasks.  Only for QVEC-cca the order is different. However, the differences between methods are small in this task. \method{} outperforms N(t) in ``doc.\ class.'', the only extrinsic task. Compared to S-ID, \method{} yields consistent improvements (except for QVEC-cca). \method{} provides higher coverage throughout all tasks.

The methods by \cite{ammar2016massively} perform well for ``doc.\ class.'', word similarity and QVEC-cca (the latter mostly because of increased dimensionality) and much worse for word translation. There are strong indications that neither of the four methods are applicable to highly multilingual corpora like PBC. ``Invariance'' considers the full
cooccurrence matrix across all languages, a matrix in the size of terabytes. In addition, word alignment
matrices would need to be stored. Both
multiCluster and multiCCA rely on bilingual dictionaries,
which are not feasible to process in the case of
PBC. multiSkip requires adding $\mathcal{O}(n^2)$ terms in the
objective function, which does not scale either.

In short: Among the methods
that are applicable to both PBC and Europarl,
\method{} performs best, followed by S-ID and N(t).

\section{Concept Quality}
\seclabel{concepts}

\begin{table}[t]
	\centering
	\scriptsize
	\def\statsep{0.2cm}
	\begin{tabular}{l@{\hspace{\statsep}}r@{\hspace{\statsep}}|r@{\hspace{\statsep}}r@{\hspace{\statsep}}r@{\hspace{\statsep}}r@{\hspace{\statsep}}r@{\hspace{\statsep}}}
		&& Mean & Median & Stddev. & Min & Max\\
		\hline
		\hline
		\multirow{2}{0.5cm}{Bible} &
		\rule{0pt}{\bigsepv}\#editions & 250 & 194 & 160 & 101 & 1530\\
		&\#tokens & 259 & 198 & 172 & 102 & 2163\\
		\hline
		\multirow{2}{1cm}{EuroParl}&
		\rule{0pt}{\bigsepv}\#editions&8&7&1&7&13\\
		&\#tokens&11&10&4&8&38\\
	\end{tabular}
	\\
	\vspace{0.2cm}
	\centering
	\scriptsize
	\def\statsep{0.2cm}
	\begin{tabular}{r@{\hspace{\statsep}}|r@{\hspace{\statsep}}r@{\hspace{\statsep}}r@{\hspace{\statsep}}r@{\hspace{\statsep}}r@{\hspace{\statsep}}}
		& \#Concepts & Cov. & Cov. (relev.) & Cov. (rare) & Cov. (freq.)\\
		\hline
		\hline
		Bible &119,026 & 0.43 & 0.56 & 0.19 & 0.85 \\
		EuroParl & 6,208,134 & 0.24 & 0.66 & 0.14 & 0.61 \\
	\end{tabular}
	\caption{Top:
		Descriptive
		statistics of
		concept
		size. Bottom:
		Coverage of
		concepts. We
		report the
		percentage of
		vocabulary in
		English (\KJV{} for PBC) that is covered by the concepts. For ``relev.'' we consider only words above the \MINC{} threshold. To examine frequent and rare words we report the coverage on the bottom/top decile based on word frequency.}
	\tablabel{conceptcoverage}
	\tablabel{conceptstats}
\end{table}

\begin{figure}[t]
	\centering
	\includegraphics[width=0.90\linewidth]{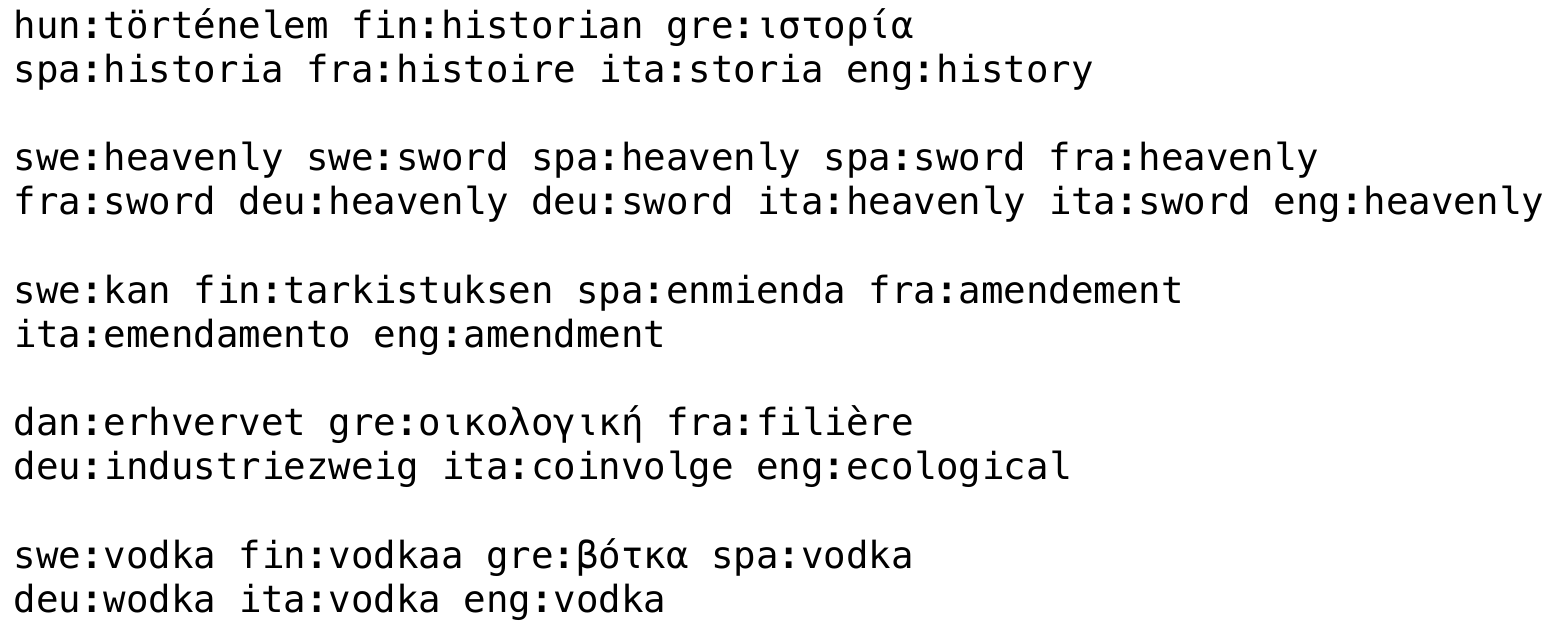}
	\caption{Five randomly sampled concepts from EuroParl. Quality is generally high. The second example is a video game consisting of two words.}
	\figlabel{conceptex}
\end{figure}

\tabref{conceptstats} reports the size of a typical concept and concept coverage with respect
to the vocabulary (English). The largest concept contains 2163 tokens across 1530 editions describing the 2-gram ``Simon Peter''.
Frequent words tend to be better covered. However, almost 20\% of really rare words (i.e., hapax legomena) are contained in concepts.
In \figref{conceptex} we show five randomly sampled concepts
from EuroParl.

\section{Related Work}
\seclabel{related}

We cluster prior work for
\textbf{multilingual embedding learning}  
for parallel corpora into three groups. Our focus is on methods which are applicable to both PBC and EuroParl.
\textbf{1)}
follows the basic idea of projecting monolingual spaces into a unified multilingual space using (linear) transformations. We use \cite{mikolov2013exploiting}
together with \cite{duong2017multilingual} in our baseline LINEAR. \citet{zou2013bilingual},
\citet{xiao2014distributed} and
\citet{faruqui2014improving} 
use similar approaches (e.g., by computing the transformation using CCA). It has been shown that computing the transformation using discriminator neural networks works well, even in a completely unsupervised setting. See, e.g., \cite{vulic2012detecting,lample2018word,chen2018unsupervised,artetxe2018robust}. We used \cite{lample2018word} as the baseline MUSE.
\textbf{2)}
is true multilingual embedding learning: it  integrates multilingual information in
the objective of  embedding
learning. \citet{klementiev2012inducing} and \citet{gouws2015bilbowa} add a word alignment based term.
\citet{luong2015bilingual} introduce BiSkip as a bilingual
extension of word2vec.
For $n$ editions, including
$\mathcal{O}(n^2)$ bilingual terms does not scale. Thus this line of work is not applicable to PBC. A slightly different objective function expresses that 
representation of aligned sentences should be
similar.
Approaches based on neural networks are  
\cite{hermann2013multilingual} (BiCVM),  
\cite{ap2014autoencoder} (autoencoders) and \cite{soyer2014leveraging}. Again, we argue that neural network based approaches do not work for the low-resource setting of PBC.
\textbf{3)}
creates multilingual corpora and uses monolingual embedding learners. 
A successful approach is \cite{levy2017strong}'s sentence ID (S-ID). \citet{vulic2015bilingual} create pseudocorpora by merging words from
multiple languages into a single corpus. \citet{dufter2018embedding}
found this method to perform poorly on PBC. \citet{sogaard2015inverted} learn a
space by factorizing an interlingual matrix based on
Wikipedia concepts. word2vec is roughly equivalent
to matrix factorization \cite{levy14neural2}, so this work fits this group.

With the raise of pretrained language models, methods to obtain \textbf{multilingual contextual representations} have been proposed \cite{conneau2019cross}. We focus on creating static word embedings which are computationally much more efficient at the cost of lower performance.

Much research has been dedicated to identifying \textbf{multilingual concepts}. BabelNet 
\cite{NavigliPonzetto:12aij} leverages existing resources
(mostly manual annotations), including Wikipedia, using information extraction methods. BabelNet could
be used to learn concept based embeddings, but it
covers only 284 languages and thus cannot be
applied to all PBC languages. Other work induces concepts within a dictionary graph \cite{ammar2016massively,dufter2018embedding}, with alignment algorithms \cite{ostling2014bayesian}, or by means of sampling \cite{lardilleux2009sampling}. We used sampling based concept induction in this work, as it scales easily for 1000+ languages.

\begin{table}
	\scriptsize
	\def\sumsep{0.1cm}
	\begin{tabular}{r@{\hspace{\sumsep}}||c@{\hspace{\sumsep}}c@{\hspace{\sumsep}}|c@{\hspace{\sumsep}}c@{\hspace{\sumsep}}c@{\hspace{\sumsep}}c@{\hspace{\sumsep}}|c@{\hspace{\sumsep}}c@{\hspace{\sumsep}}}
		& \multicolumn{2}{c}{PBC} & \multicolumn{4}{c}{EuroParl} & & \\
		&  & & Doc. & Word & QVEC & Word  & Mean & Mean \\
		& RTT & Sent. & Class. & Sim. & -cca & Trans. & Rank &Perf. \\
		\hline
		S-ID & 3 & 3 & 2 & 2 & 2 & 2 &  2.33 & 53.91 \\
		N(t) &\textbf{1}&\textbf{1}& 3 & 3 &\textbf{1}& 3 & 2.00 & 55.87 \\
		Co+Co & 2 & 2 &\textbf{1}&\textbf{1}& 3 &\textbf{1}& \textbf{1.67} & \textbf{ 56.31} 
	\end{tabular}
	\centering
	\caption{Performance overview: we show the rank among N(t), S-ID and \method{} across all tasks. For RTT and Sent. the overall performance is the mean over all task versions (S1, R1, S4, S16 and Pos., Neg.). }
	\tablabel{summaryall}
\end{table}

\section{Summary}
\seclabel{summary}
We proposed \method{}, to the best of our knowledge the first method that learns embeddings \emph{jointly
	from concept and context  information}. 
We showed that
\method{} performs
well across two very
different corpora
and a wide range of
tasks. 
Among the three
high-performing methods applicable to both PBC and EuroParl \method{} performs best (see \tabref{summaryall}).
Two other advantages of \method{} are that it is a simple
method (compared to more complex methods like MUSE) and
scalable to 1000s of languages. In summary,
\method{} is a simple, strong and scalable method that
is well suited for a wide range of application scenarios.

We gratefully \textbf{acknowledge} funding for this work by
the European Research
Council (ERC \#740516) and by
Zentrum Digitalisierung.Bayern (ZD.B), the digital technology
initiative of the State of Bavaria.

\bibliography{stronger}

\begin{thebibliography}{39}
\expandafter\ifx\csname natexlab\endcsname\relax\def\natexlab#1{#1}\fi

\bibitem[{Ammar et~al.(2016)Ammar, Mulcaire, Tsvetkov, Lample, Dyer, and
  Smith}]{ammar2016massively}
Waleed Ammar, George Mulcaire, Yulia Tsvetkov, Guillaume Lample, Chris Dyer,
  and Noah~A Smith. 2016.
\newblock Massively multilingual word embeddings.
\newblock \emph{arXiv preprint arXiv:1602.01925}.

\bibitem[{Artetxe et~al.(2018)Artetxe, Labaka, and Agirre}]{artetxe2018robust}
Mikel Artetxe, Gorka Labaka, and Eneko Agirre. 2018.
\newblock A robust self-learning method for fully unsupervised cross-lingual
  mappings of word embeddings.
\newblock In \emph{Proceedings of the 56th Annual Meeting of the Association
  for Computational Linguistics (Volume 1: Long Papers)}, pages 789--798.

\bibitem[{Chen and Cardie(2018)}]{chen2018unsupervised}
Xilun Chen and Claire Cardie. 2018.
\newblock Unsupervised multilingual word embeddings.
\newblock In \emph{Proceedings of the 2018 Conference on Empirical Methods in
  Natural Language Processing}.

\bibitem[{Conneau and Lample(2019)}]{conneau2019cross}
Alexis Conneau and Guillaume Lample. 2019.
\newblock Cross-lingual language model pretraining.
\newblock In \emph{Advances in Neural Information Processing Systems}, pages
  7057--7067.

\bibitem[{Dufter et~al.(2018)Dufter, Zhao, Schmitt, Fraser, and
  Sch{\"u}tze}]{dufter2018embedding}
Philipp Dufter, Mengjie Zhao, Martin Schmitt, Alexander Fraser, and Hinrich
  Sch{\"u}tze. 2018.
\newblock Embedding learning through multilingual concept induction.
\newblock In \emph{Proceedings of the 56th Annual Meeting of the Association
  for Computational Linguistics}.

\bibitem[{Duong et~al.(2017)Duong, Kanayama, Ma, Bird, and
  Cohn}]{duong2017multilingual}
Long Duong, Hiroshi Kanayama, Tengfei Ma, Steven Bird, and Trevor Cohn. 2017.
\newblock Multilingual training of crosslingual word embeddings.
\newblock In \emph{Proceedings of the 15th Conference of the European Chapter
  of the Association for Computational Linguistics}.

\bibitem[{Faruqui and Dyer(2014)}]{faruqui2014improving}
Manaal Faruqui and Chris Dyer. 2014.
\newblock Improving vector space word representations using multilingual
  correlation.
\newblock In \emph{Proceedings of the 14th Conference of the European Chapter
  of the Association for Computational Linguistics}.

\bibitem[{Gouws et~al.(2015)Gouws, Bengio, and Corrado}]{gouws2015bilbowa}
Stephan Gouws, Yoshua Bengio, and Greg Corrado. 2015.
\newblock Bilbowa: fast bilingual distributed representations without word
  alignments.
\newblock In \emph{Proceedings of the 32nd International Conference on
  International Conference on Machine Learning}.

\bibitem[{Guo et~al.(2016)Guo, Che, Yarowsky, Wang, and Liu}]{guo16transfer}
Jiang Guo, Wanxiang Che, David Yarowsky, Haifeng Wang, and Ting Liu. 2016.
\newblock A representation learning framework for multi-source transfer
  parsing.
\newblock In \emph{Proceedings of the 30th AAAI Conference on Artificial
  Intelligence}.

\bibitem[{Hermann and Blunsom(2014{\natexlab{a}})}]{hermann2013multilingual}
Karl~Moritz Hermann and Phil Blunsom. 2014{\natexlab{a}}.
\newblock Multilingual distributed representations without word alignment.
\newblock In \emph{Proceedings of the 2014 International Conference on Learning
  Representations}.

\bibitem[{Hermann and Blunsom(2014{\natexlab{b}})}]{hermann2014multilingual}
Karl~Moritz Hermann and Phil Blunsom. 2014{\natexlab{b}}.
\newblock Multilingual models for compositional distributed semantics.
\newblock In \emph{Proceedings of the 52nd Annual Meeting of the Association
  for Computational Linguistics}.

\bibitem[{Klementiev et~al.(2012)Klementiev, Titov, and
  Bhattarai}]{klementiev2012inducing}
Alexandre Klementiev, Ivan Titov, and Binod Bhattarai. 2012.
\newblock Inducing crosslingual distributed representations of words.
\newblock In \emph{Proceedings of the 24th International Conference on
  Computational Linguistics}.

\bibitem[{Lample et~al.(2018)Lample, Conneau, Ranzato, Denoyer, and
  Jégou}]{lample2018word}
Guillaume Lample, Alexis Conneau, Marc'Aurelio Ranzato, Ludovic Denoyer, and
  Hervé Jégou. 2018.
\newblock Word translation without parallel data.
\newblock In \emph{Proceedings of the 2018 International Conference on Learning
  Representations}.

\bibitem[{Lardilleux and Lepage(2009)}]{lardilleux2009sampling}
Adrien Lardilleux and Yves Lepage. 2009.
\newblock Sampling-based multilingual alignment.
\newblock In \emph{Proceedings of 7th Conference on Recent Advances in Natural
  Language Processing}.

\bibitem[{Levy and Goldberg(2014)}]{levy14neural2}
Omer Levy and Yoav Goldberg. 2014.
\newblock Neural word embedding as implicit matrix factorization.
\newblock In \emph{Advances in Neural Information Processing Systems 27: Annual
  Conference on Neural Information Processing Systems 2014}.

\bibitem[{Levy et~al.(2017)Levy, S{\o}gaard, and Goldberg}]{levy2017strong}
Omer Levy, Anders S{\o}gaard, and Yoav Goldberg. 2017.
\newblock A strong baseline for learning cross-lingual word embeddings from
  sentence alignments.
\newblock In \emph{Proceedings of the 15th Conference of the European Chapter
  of the Association for Computational Linguistics}.

\bibitem[{Li and Jurafsky(2015)}]{li15multisense}
Jiwei Li and Dan Jurafsky. 2015.
\newblock Do multi-sense embeddings improve natural language understanding?
\newblock In \emph{Proceedings of the 2015 Conference on Empirical Methods in
  Natural Language Processing}.

\bibitem[{Luong et~al.(2015)Luong, Pham, and Manning}]{luong2015bilingual}
Thang Luong, Hieu Pham, and Christopher~D Manning. 2015.
\newblock Bilingual word representations with monolingual quality in mind.
\newblock In \emph{Proceedings of the 1st Workshop on Vector Space Modeling for
  Natural Language Processing}.

\bibitem[{Mayer and Cysouw(2014)}]{mayer2014creating}
Thomas Mayer and Michael Cysouw. 2014.
\newblock Creating a massively parallel bible corpus.
\newblock In \emph{Proceedings of the 9th International Conference on Language
  Resources and Evaluation}.

\bibitem[{McDonald et~al.(2011)McDonald, Petrov, and
  Hall}]{mcdonald11delexicalized}
Ryan~T. McDonald, Slav Petrov, and Keith~B. Hall. 2011.
\newblock Multi-source transfer of delexicalized dependency parsers.
\newblock In \emph{Proceedings of the 2011 Conference on Empirical Methods in
  Natural Language Processing}.

\bibitem[{Mikolov et~al.(2013{\natexlab{a}})Mikolov, Chen, Corrado, and
  Dean}]{mikolov2013efficient}
Tomas Mikolov, Kai Chen, Greg Corrado, and Jeffrey Dean. 2013{\natexlab{a}}.
\newblock Efficient estimation of word representations in vector space.
\newblock \emph{arXiv preprint arXiv:1301.3781}.

\bibitem[{Mikolov et~al.(2013{\natexlab{b}})Mikolov, Le, and
  Sutskever}]{mikolov2013exploiting}
Tomas Mikolov, Quoc~V Le, and Ilya Sutskever. 2013{\natexlab{b}}.
\newblock Exploiting similarities among languages for machine translation.
\newblock \emph{arXiv preprint arXiv:1309.4168}.

\bibitem[{Navigli and Ponzetto(2012)}]{NavigliPonzetto:12aij}
Roberto Navigli and Simone~Paolo Ponzetto. 2012.
\newblock {B}abel{N}et: {T}he automatic construction, evaluation and
  application of a wide-coverage multilingual semantic network.
\newblock \emph{Artificial Intelligence}.

\bibitem[{{\"O}stling(2014)}]{ostling2014bayesian}
Robert {\"O}stling. 2014.
\newblock Bayesian word alignment for massively parallel texts.
\newblock In \emph{Proceedings of the 14th Conference of the European Chapter
  of the Association for Computational Linguistics}.

\bibitem[{Penrose(1956)}]{penrose1956best}
Roger Penrose. 1956.
\newblock On best approximate solutions of linear matrix equations.
\newblock In \emph{Mathematical Proceedings of the Cambridge Philosophical
  Society}.

\bibitem[{Ruder et~al.(2019)Ruder, Vuli{\'c}, and S{\o}gaard}]{ruder2019survey}
Sebastian Ruder, Ivan Vuli{\'c}, and Anders S{\o}gaard. 2019.
\newblock A survey of cross-lingual word embedding models.
\newblock \emph{Journal of Artificial Intelligence Research}, 65:569--631.

\bibitem[{{Sarath Chandar} et~al.(2014){Sarath Chandar}, Lauly, Larochelle,
  Khapra, Ravindran, Raykar, and Saha}]{ap2014autoencoder}
AP~{Sarath Chandar}, Stanislas Lauly, Hugo Larochelle, Mitesh~M. Khapra,
  Balaraman Ravindran, Vikas~C. Raykar, and Amrita Saha. 2014.
\newblock An autoencoder approach to learning bilingual word representations.
\newblock In \emph{Proceedings of the 2014 Annual Conference on Neural
  Information Processing Systems}.

\bibitem[{S{\o}gaard et~al.(2015)S{\o}gaard, Agi{\'c}, Alonso, Plank, Bohnet,
  and Johannsen}]{sogaard2015inverted}
Anders S{\o}gaard, {\v{Z}}eljko Agi{\'c}, H{\'e}ctor~Mart{\'\i}nez Alonso,
  Barbara Plank, Bernd Bohnet, and Anders Johannsen. 2015.
\newblock Inverted indexing for cross-lingual nlp.
\newblock In \emph{The 53rd Annual Meeting of the Association for Computational
  Linguistics and the 7th International Joint Conference of the Asian
  Federation of Natural Language Processing}.

\bibitem[{Soyer et~al.(2014)Soyer, Stenetorp, and Aizawa}]{soyer2014leveraging}
Hubert Soyer, Pontus Stenetorp, and Akiko Aizawa. 2014.
\newblock Leveraging monolingual data for crosslingual compositional word
  representations.
\newblock In \emph{Proceedings of the 2015 International Conference on Learning
  Representations}.

\bibitem[{Swadesh(1946)}]{swadesh1971south}
Morris Swadesh. 1946.
\newblock South {G}reenlandic ({E}skimo).
\newblock In Cornelius Osgood, editor, \emph{Linguistic Structures of Native
  America}. Viking Fund Inc. (Johnson Reprint Corp.), New York.

\bibitem[{Tsvetkov et~al.(2014)Tsvetkov, Boytsov, Gershman, Nyberg, and
  Dyer}]{tsvetkov14metaphor}
Yulia Tsvetkov, Leonid Boytsov, Anatole Gershman, Eric Nyberg, and Chris Dyer.
  2014.
\newblock Metaphor detection with cross-lingual model transfer.
\newblock In \emph{Proceedings of the 52nd Annual Meeting of the Association
  for Computational Linguistics}.

\bibitem[{Tsvetkov et~al.(2015)Tsvetkov, Faruqui, Ling, Lample, and
  Dyer}]{tsvetkov2015evaluation}
Yulia Tsvetkov, Manaal Faruqui, Wang Ling, Guillaume Lample, and Chris Dyer.
  2015.
\newblock Evaluation of word vector representations by subspace alignment.
\newblock In \emph{Proceedings of the 2015 Conference on Empirical Methods in
  Natural Language Processing}.

\bibitem[{Upadhyay et~al.(2016)Upadhyay, Faruqui, Dyer, and
  Roth}]{upadhyay2016cross}
Shyam Upadhyay, Manaal Faruqui, Chris Dyer, and Dan Roth. 2016.
\newblock Cross-lingual models of word embeddings: An empirical comparison.
\newblock In \emph{Proceedings of the 54th Annual Meeting of the Association
  for Computational Linguistics}.

\bibitem[{Vuli{\'c} and Moens(2012)}]{vulic2012detecting}
Ivan Vuli{\'c} and Marie-Francine Moens. 2012.
\newblock Detecting highly confident word translations from comparable corpora
  without any prior knowledge.
\newblock In \emph{Proceedings of the 13th Conference of the European Chapter
  of the Association for Computational Linguistics}.

\bibitem[{Vuli{\'c} and Moens(2015)}]{vulic2015bilingual}
Ivan Vuli{\'c} and Marie-Francine Moens. 2015.
\newblock Bilingual word embeddings from non-parallel document-aligned data
  applied to bilingual lexicon induction.
\newblock In \emph{Proceedings of the 53rd Annual Meeting of the Association
  for Computational Linguistics and the 7th International Joint Conference on
  Natural Language Processing}.

\bibitem[{Xiao and Guo(2014)}]{xiao2014distributed}
Min Xiao and Yuhong Guo. 2014.
\newblock Distributed word representation learning for cross-lingual dependency
  parsing.
\newblock In \emph{Proceedings of the 18th Conference on Computational Natural
  Language Learning}.

\bibitem[{Yin and Shen(2018)}]{yin2018dimensionality}
Zi~Yin and Yuanyuan Shen. 2018.
\newblock On the dimensionality of word embedding.
\newblock In \emph{Advances in Neural Information Processing Systems}.

\bibitem[{Zeman and Resnik(2008)}]{zeman08crosslanguage}
Daniel Zeman and Philip Resnik. 2008.
\newblock Cross-language parser adaptation between related languages.
\newblock In \emph{Proceedings of the 3rd International Joint Conference on
  Natural Language Processing}.

\bibitem[{Zou et~al.(2013)Zou, Socher, Cer, and Manning}]{zou2013bilingual}
Will~Y Zou, Richard Socher, Daniel Cer, and Christopher~D Manning. 2013.
\newblock Bilingual word embeddings for phrase-based machine translation.
\newblock In \emph{Proceedings of the 2013 Conference on Empirical Methods in
  Natural Language Processing}.

\end{thebibliography}
\bibliographystyle{acl_natbib}

\end{document}